%% file: AdaTrip/main.tex
\documentclass[conference]{IEEEtran}

\usepackage{cite}
\usepackage{amsmath,amssymb,amsfonts}
\usepackage{graphicx}
\usepackage{textcomp}
\usepackage{xcolor}
\def\BibTeX{{\rm B\kern-.05em{\sc i\kern-.025em b}\kern-.08em
    T\kern-.1667em\lower.7ex\hbox{E}\kern-.125emX}}

\usepackage{booktabs}
\usepackage{multirow}
\usepackage{hyperref}

\usepackage{algorithm}
\usepackage{algpseudocode}

\usepackage[normalem]{ulem}

\newcommand{\model}{\texttt{AdaTrip}}

\begin{document}

\makeatletter
\newcommand{\linebreakand}{%
  \end{@IEEEauthorhalign}
  \hfill\mbox{}\par
  \mbox{}\hfill\begin{@IEEEauthorhalign}
}

\def\ps@IEEEtitlepagestyle{%
  \def\@oddfoot{\mycopyrightnotice}%
  \def\@evenfoot{}%
}
\def\mycopyrightnotice{%
  {\footnotesize \begin{minipage}{\textwidth}\ 
  This manuscript has been authored by UT-Battelle LLC, under contract DE-AC05-00OR22725 with the US Department of Energy (DOE). The US government retains and the publisher, by accepting the article for publication, acknowledges that the US government retains a nonexclusive, paid-up, irrevocable, worldwide license to publish or reproduce the published form of this manuscript, or allow others to do so, for US government purposes. DOE will provide public access to these results of federally sponsored research in accordance with the DOE Public Access Plan (http://energy.gov/downloads/doe-public-access-plan).
\end{minipage}\hfill}
  \gdef\mycopyrightnotice{}
}

\makeatother

\title{Adaptive Graph Learning with Transformer for Multi-Reservoir Inflow Prediction}

\author{Anonymous}

\author{
\IEEEauthorblockN{
Pengfei Hu,\textsuperscript{1,2}\thanks{Work done while interning at Oak Ridge National Laboratory}
Ming Fan,\textsuperscript{2}
Xiaoxue Han,\textsuperscript{1}
Chang Lu,\textsuperscript{1}
Wei Zhang,\textsuperscript{2}
Hyun Kang,\textsuperscript{2}
Yue Ning,\textsuperscript{1}
Dan Lu\textsuperscript{2}
}\\
\IEEEauthorblockA{\textsuperscript{1}Stevens Institute of Technology, Hoboken, NJ, USA \\
\{phu9, xhan26, yue.ning\}@stevens.edu, luchang.cs@gmail.com}
\IEEEauthorblockA{\textsuperscript{2}Oak Ridge National Laboratory, Oak Ridge, TN, USA \\
\{fanm, zhangw3, kangh, lud1\}@ornl.gov}
}

\maketitle

\begin{abstract}
Reservoir inflow prediction is crucial for water resource management, yet existing approaches mainly focus on single-reservoir models that ignore spatial dependencies among interconnected reservoirs. 
We introduce \model{} as an adaptive, time-varying graph learning framework for multi-reservoir inflow forecasting. \model{} constructs dynamic graphs where reservoirs are nodes with directed edges reflecting hydrological connections, employing attention mechanisms to automatically identify crucial spatial and temporal dependencies. Evaluation on thirty reservoirs in the Upper Colorado River Basin demonstrates superiority over existing baselines, with improved performance for reservoirs with limited records through parameter sharing. Additionally, \model{} provides interpretable attention maps at edge and time-step levels, offering insights into hydrological controls to support operational decision-making. 
Our code is available at \href{https://github.com/humphreyhuu/AdaTrip}{https://github.com/humphreyhuu/AdaTrip}.
\end{abstract}

\begin{IEEEkeywords}
Graph Neural Network, Reservoir Inflow, Hydrology, Adaptive Graph, Explainable Machine Learning.
\end{IEEEkeywords}

\section{Introduction}
\label{sec:intro}
\input{sections/intro}

\section{Related Work}
\input{sections/related_work}

\begin{figure*}[t]
    \centering
    \includegraphics[width=\textwidth]{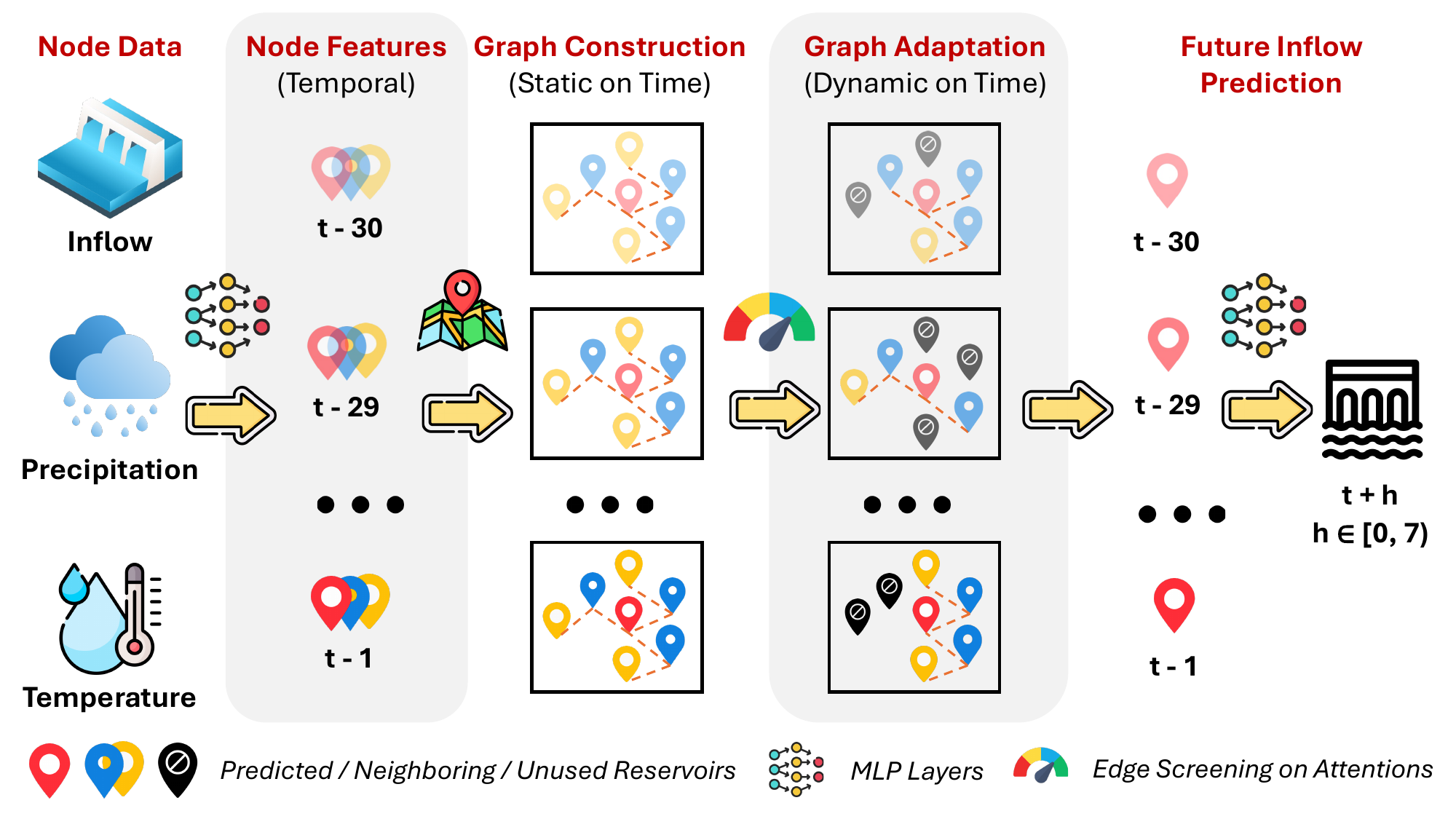}
    \caption{Overall Framework of \model{}. Starting with historical reservoir inflow observations $\{\mathbf{x}_{t}\}_{t=1}^{T}$, precipitation, and temperature, a feature extractor $\mathcal{F}_{\theta}$ maps them to embeddings $\mathbf{h}_{t}=\operatorname{MLP}(\mathbf{x}_{t})$; based on pairwise reservoir distances $d_{ij}$, an initial static graph $\mathcal{G}_{0}\!=\!(\mathcal{V},\mathcal{E},\mathbf{A})$ with adjacency $a_{ij}$ is formed; a graph attention network $\mathcal{G}_{\lambda}$ assigns time‑varying edge weights and removes edges with $\alpha_{ij,t}<\tau$ to obtain the adaptive graph $\mathcal{G}_{t}$; an encoder–decoder transformer $(\mathcal{T}_{e},\mathcal{T}_{d})$ summarizes $\mathbf{M}_{i}$ into latent states $\mathbf{z}_{i}$, and a linear layer yields $k\!\in\![1,7]$‑day ahead inflow predictions $\hat{y}_{i,t+k}$ for every reservoir.}
    \label{fig:framework}
\end{figure*}

\section{Preliminary}
\input{tabs/notation}
\input{sections/preliminary}

\section{Methodology}
\input{sections/method}

\section{Experiment}

\subsection{Data}

We apply \model{} to reservoirs in the Upper Colorado River Basin, which spans Colorado, New Mexico, Utah, and Wyoming, supplying fresh water primarily through winter snowpack and spring runoff to nearly 40 million people while supporting hydropower generation, flood control, irrigation, and recreation ~\cite{fan2022identifying}. 
Here we analyze 30 reservoirs (as shown in Table \ref{tab:stats}) selected based on complete hydrological records of daily inflow data with minimal gaps (less than ten days) obtained from the U.S. Bureau of Reclamation water operation archive ~\cite{fan2023investigation}. 
We also utilize the AN81d dataset derived from the Parameter-elevation Regressions on Independent Slopes Model (PRISM), which provides high-resolution spatial coverage at 4 km resolution (approximately 0.04 degrees) across the basin ~\cite{daly2013prism}. These reservoirs vary in elevation, storage capacity, and operational purposes, with data analysis covering a 13-year period determined by the shortest available record to ensure consistency across all selected reservoirs, enabling comprehensive assessment of reservoir performance under varying hydrological and meteorological conditions.


\input{tabs/data_stats}

\subsection{Evaluation Metrics}

The model’s prediction accuracy is assessed using the Nash--Sutcliffe Efficiency (NSE) metric. NSE quantifies the consistency between the predicted reservoir inflow and its observed values, taking into account both the variations in predicted inflow and the cumulative biases between the predicted and observed values, as in Eq. (\ref{eq:nse}),
\begin{equation}
\text{NSE}(\hat{y}_i, {y}_i) = 1 - \frac{\sum_{i=1}^{n} (\hat{y}_i - y_i)^2}{\sum_{i=1}^{n} (y_i - \bar{y}_i)^2}
\label{eq:nse}
\end{equation}
where $y_i$ and $\hat{y}_i$ are the observations and predicted reservoir inflow, respectively, $\bar{y}_i$ refers to the mean value of the observations, and $n$ represents the total number of observations. The range of the NSE is $(-\infty, 1]$, where a value of 1 denotes the best prediction performance. 

\input{tabs/main_results}

\begin{figure*}[t]
    \centering
    \includegraphics[width=0.8\textwidth]{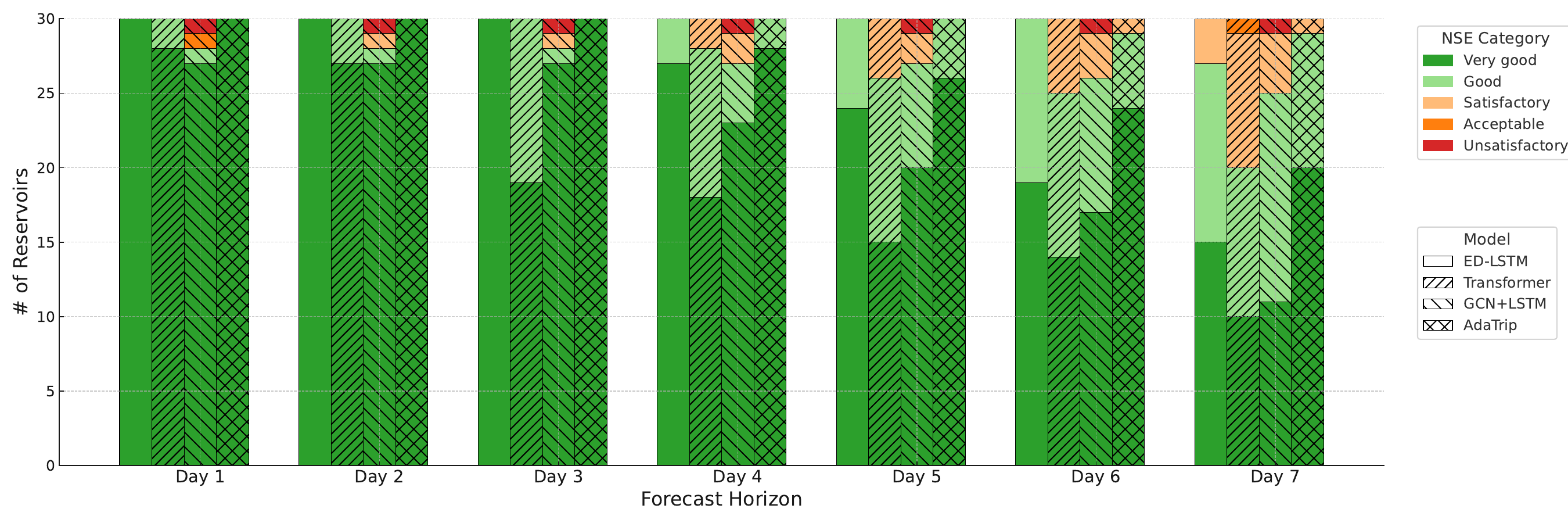}
    \caption{Comparison of the prediction performance between the ED--LSTM, Transformer, GCN+LSTM, and \model{} models for the 30-day forecasting horizons using the NSE metric
    (Very good: \( \text{NSE} > 0.75 \); Good: \(0.65 < \text{NSE} \le 0.75\); Satisfactory: \(0.5 < \text{NSE} \le 0.65\); Acceptable: \(0.4 < \text{NSE} \le 0.5\); and Unsatisfactory: \( \text{NSE} \le 0.4 \)).  
    Results are shown for all three baselines and our \model{} across the seven-day forecast horizon.}
    \label{fig:res_horizon_histogram}
\end{figure*}

\subsection{Baselines}

To quantify the contribution of dynamic graph learning we compare \model{} against three representative neural architectures that capture temporal patterns but differ in how they learn spatial-temporal patterns. 

\begin{itemize}
    \item \textbf{ED-LSTM~\cite{fan2023explainable}.} A sequence-to-sequence LSTM network with an encoder layer and a decoder layer. Each reservoir is treated as an independent series: the past inflow, temperature, and precipitation vectors are packed into a sequence, and the decoder generates the next inflow values. The model shares parameters over reservoirs but has no explicit spatial interaction.
    
    \item \textbf{Transformer~\cite{vaswani2017attention}.} A Transformer that replaces recurrence with 8 multi-head self-attention modules.  The daily feature vectors of each reservoir pass through 4 encoder blocks; a decoder with the same depth produces the seven-day forecast. Similarly, it also ignores reservoir connectivity like ED-LSTM.
    
    \item \textbf{GCN\,+\,LSTM~\cite{yu2017spatio}.} Node features are firstly processed by single graph convolution layer, and the adjusted embeddings after GCN are then fed to an ED-LSTM module. Note that the spatial mixing is fixed and uniform due to pre-defined weights in the adjacent matrix. 
\end{itemize}

\subsection{Implementation Details}

We randomly initialize all embeddings and model parameters.
The embedding sizes $d,m,r$ for $\mathbf{h}_{i,t}, \tilde{\mathbf{h}}_{i,t}, \mathbf{z}_i$ in $\mathcal{F}_\theta, \mathcal{G}_\lambda,\mathcal{T}_\omega$ are 128, 128, and 64, respectively. 
We use two graph attention convolution layers (4 heads and each for 32 hidden units) as where the hidden unit number is 128 with $K=4$ heads, and $\mathcal{G}_\lambda$ learn edges involving self loops $\delta_{i,i}$ with $i\leq30$ for all reservoirs. 
For graph construction, we initialize daily graphs $G_t$ and adjacent matrices $\mathbf{A}_t$ by filtering out $k=2$ nearest neighbors as $\mathcal{N}_i$ for each nodes. 
The iterative graph adaptation is triggered every $m=4$ epochs: edges whose mean attention falls below $\tau=0.3$ are masked.
For temporal predictor, we utilize the adjusted encoder-decoder transformer as $\mathcal{T}_\omega(\cdot)$ with 2 layers and 4 heads on each layer, and the feed-forward width is 256. 
Pre-training runs for 5 epochs with a contrastive and supervised loss weights schedule of 4:1.
In addition, for the training phase, we add Dropout on $\mathcal{G}_\lambda(\cdot)$ and $\mathcal{T}_\omega(\cdot)$ layers, and dropout rates are $0.2$ throughout. 
For the pre-training and fine-tuning phase of our task, we use the ReLU as activation functions in the feature extractor $\mathcal{F}_\theta(\cdot)$, and loss function is mean square error.

We use 5 epochs for pre-training reservoir daily embeddings.
For our supervised learning, we use 10 epochs for all baselines and our proposed model, which is equivalent to 300 epochs for single-reservoir training.
The initial learning rate is 1e-3 and decays by 0.5 every single epoch, and the batch sizes of all models are 4.
The Adam optimizer is uniformly used in both pre-training and training phase. 
All programs are implemented using Python 3.10.17 and Pytorch 2.5.1 with ROCM 6.2.4 on a computation node with 3 AMD 64-core EPYC CPU, 512 GB RAM, and eight AMD MI250X GCD with 64 GB HBM. Addition software stack for GNNs is PyTorch Geometric 2.6.

\subsection{Main Results}

\subsubsection{Overall performance}

Table~\ref{tab:overall_results} summarises the mean NSE score across those 30 reservoirs and the seven-day forecast horizons. 
\model{} attains the highest overall NSE of $91.45\%$, surpassing the three baseline methods by approximately $2.0-3.5\%$. 
The performance gap is modest for the one–step forecast (Day1) but widens markedly from Day5 onward. 
Specifically, \model{} sustains NSE values of \(88.03\%\) on Day6 and \(86.00\%\) on Day 7, whereas the best baseline declines to \(85.27\%\) and \(83.63\%\), respectively. 
Among the baselines, ED‑LSTM~\cite{fan2023explainable} performs best, underscoring the significance of modelling temporal dependencies. 
In contrast, the lower scores of GCN+LSTM~\cite{yu2017spatio} and Transformer~\cite{vaswani2017attention} indicate that increasing architectural complexity does not necessarily lead to better results and confirm the importance of topology‑sensitive design discussed earlier. 
Overall, the superior performance of \model{} demonstrates that combining high‑quality spatial information with temporal dynamics is a promising approach for multi‑reservoir inflow forecasts.


\subsubsection{Individual performance}

Figure~\ref{fig:res_horizon_histogram} groups forecasts for the thirty reservoirs into NSE performance categories over the seven‑day horizon. 
On Day~1 all models perform predominantly in the ``Very good'' range, with both \model{} and ED-LSTM keeping all reservoirs above the \(0.75\) threshold. 
Up to Day~4 the great majority of reservoirs' performance for all methods remain at least ``Good'', after which a clear decline emerges. 
Specifically, GCN+LSTM shows its first ``Unsatisfactory'' case on Day~5, indicating substantial variance in reservoir-level predictions. 
The Transformer model exhibits the most significant drop, with the most forecasts falling below``Good'' after Day 4. 
ED‑LSTM tracks \model{} closely, yet on Days 6 and 7 \model{} retains ``Good'' or better scores for more reservoirs.
These reservoir‑level observations support the overall NSE advantage of \model{} reported in Table~\ref{tab:overall_results}.

\begin{figure}[t]
    \centering
    \includegraphics[width=0.45\textwidth]{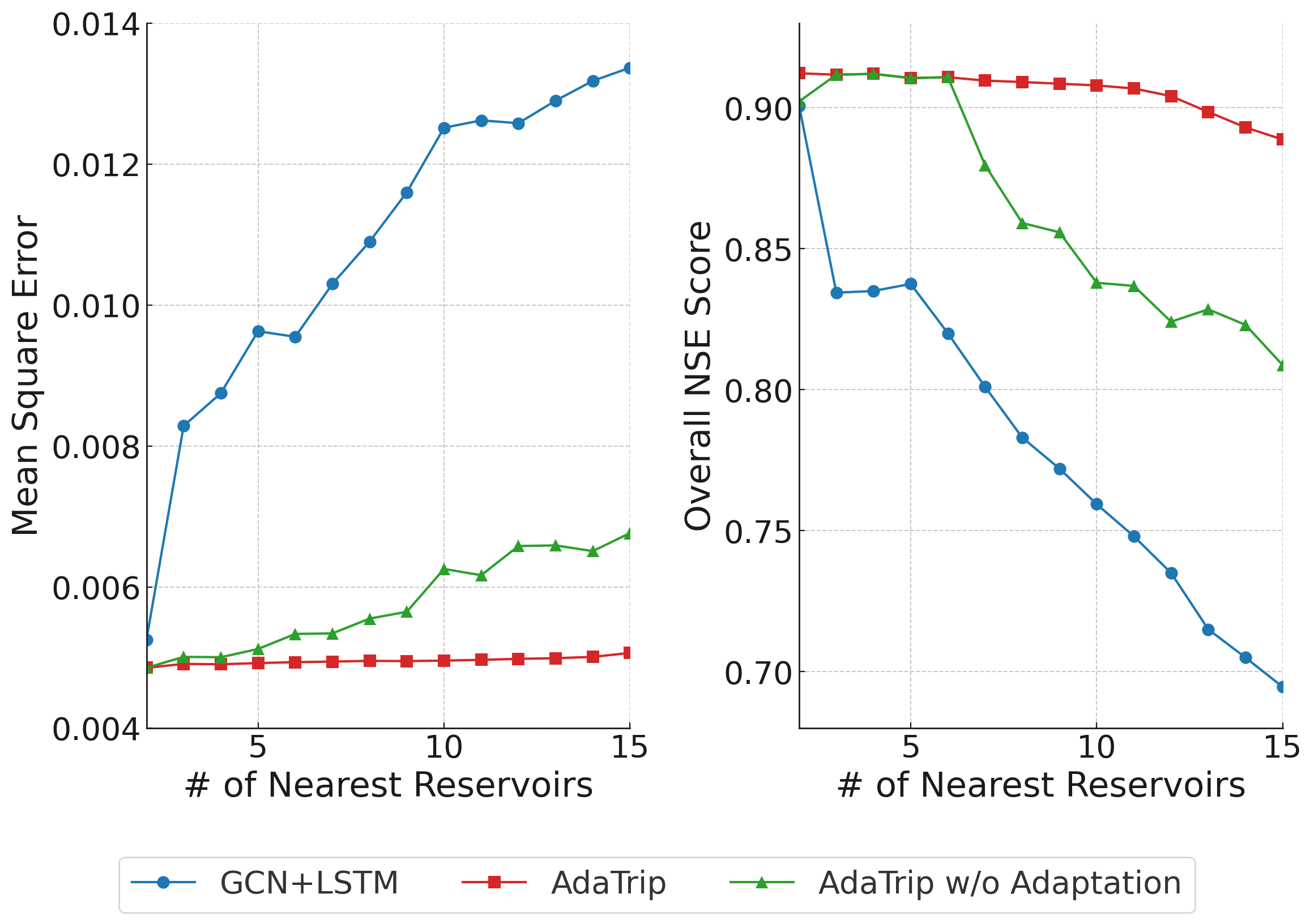}
    \caption{NSE scores and Mean Square Errors are shown across different graph constructions. We can observe that GCN+LSTM and \model{} without adaptation are more sensitive to the topology structure, which demonstrates the problem 1) in section~\ref{sec:intro} also exists in the hydrology domain.}
    \label{fig:exist_adaptation}
\end{figure}

\subsection{Effectiveness of Graph Adaptation}

Here we investigate the behavior of the graph adaptation module from two perspectives. 
First, we aim to verify whether the research questions proposed in Section~\ref{sec:intro} are indeed reflected in real-world hydrological systems. 
Second, we examine the adaptation dynamics by selecting a representative subgraph and analyzing its changes in detail.

\begin{enumerate}
    \item Figure~\ref{fig:exist_adaptation} presents the changes in NSE scores under different graph construction strategies across three experimental settings. Compared to \model{}, the GCN+LSTM baseline exhibits substantial declines in both NSE and MSE, indicating that conventional GNNs are highly sensitive to graph topology. Such sensitivity often necessitates manual graph design prior to training and lacks interpretability. Regarding temporal dynamics, the predictive performance of \model{} also degrades when the model is not allowed to adapt to evolving daily graph structures. Overall, these results demonstrate that \model{} achieves more stable and accurate predictions by automatically adapting the graph topology to fit the underlying data patterns.

    \item Figure~\ref{fig:adaptation_demo} illustrates the edge pruning procedure for a subgraph containing 6 reservoirs from the same river system. The setup matches the main study: the graph is updated every \(m=4\) training epochs. Before pruning, the graph reveals connection strength via attention weights. For instance, reservoir ECR receives 0.502 from ROC versus 0.250 from ECH, where higher weights indicate stronger predictive importance. During adaptation three edges are pruned, each connected to the downstream reservoir ECH.  Figure~\ref{fig:inflows_demo} plots the historical inflow series for the two removed edges, (ECR,,ECH) and (LCR,,ECH).
    A significant disparity exists in the inflow magnitudes—while LCR and ECH both have maximum inflows of approximately 600 cfs, ECR experiences much higher inflows reaching around 2000 cfs. This substantial difference between ECR and the similar-magnitude reservoirs (ECH and LCR) reduces the relevance of the ECR-ECH connection for reservoir inflow forecasting, leading \model{} to dynamically prune these edges during training.
\end{enumerate}

\begin{figure}[t]
    \centering
    \includegraphics[width=0.45\textwidth]{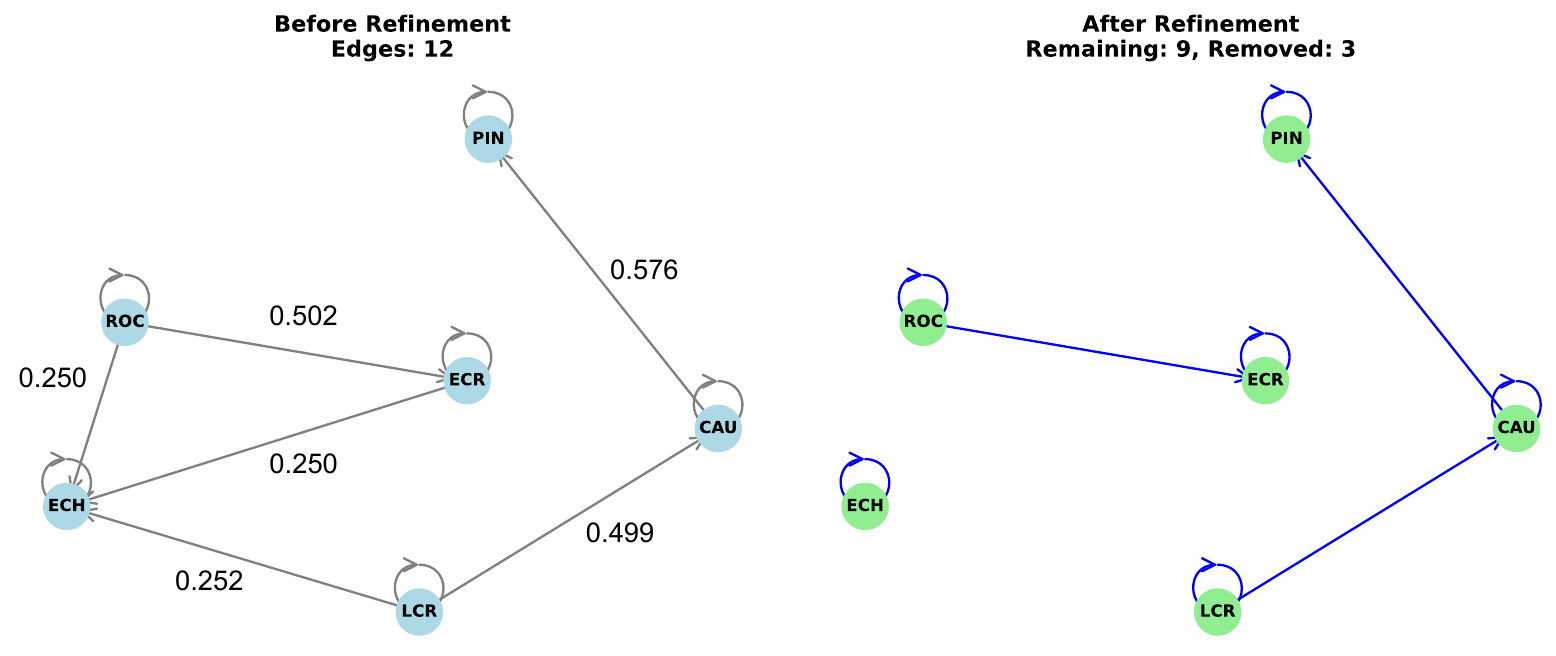}
    \caption{The demonstration of the graph adaptation on day $t=30$ after the $4$-th epoch ends in \model{}. Self-loop edges are also involved in $G_{30}$.}
    \label{fig:adaptation_demo}
\end{figure}

\begin{figure}[t]
    \centering
    \includegraphics[width=0.45\textwidth]{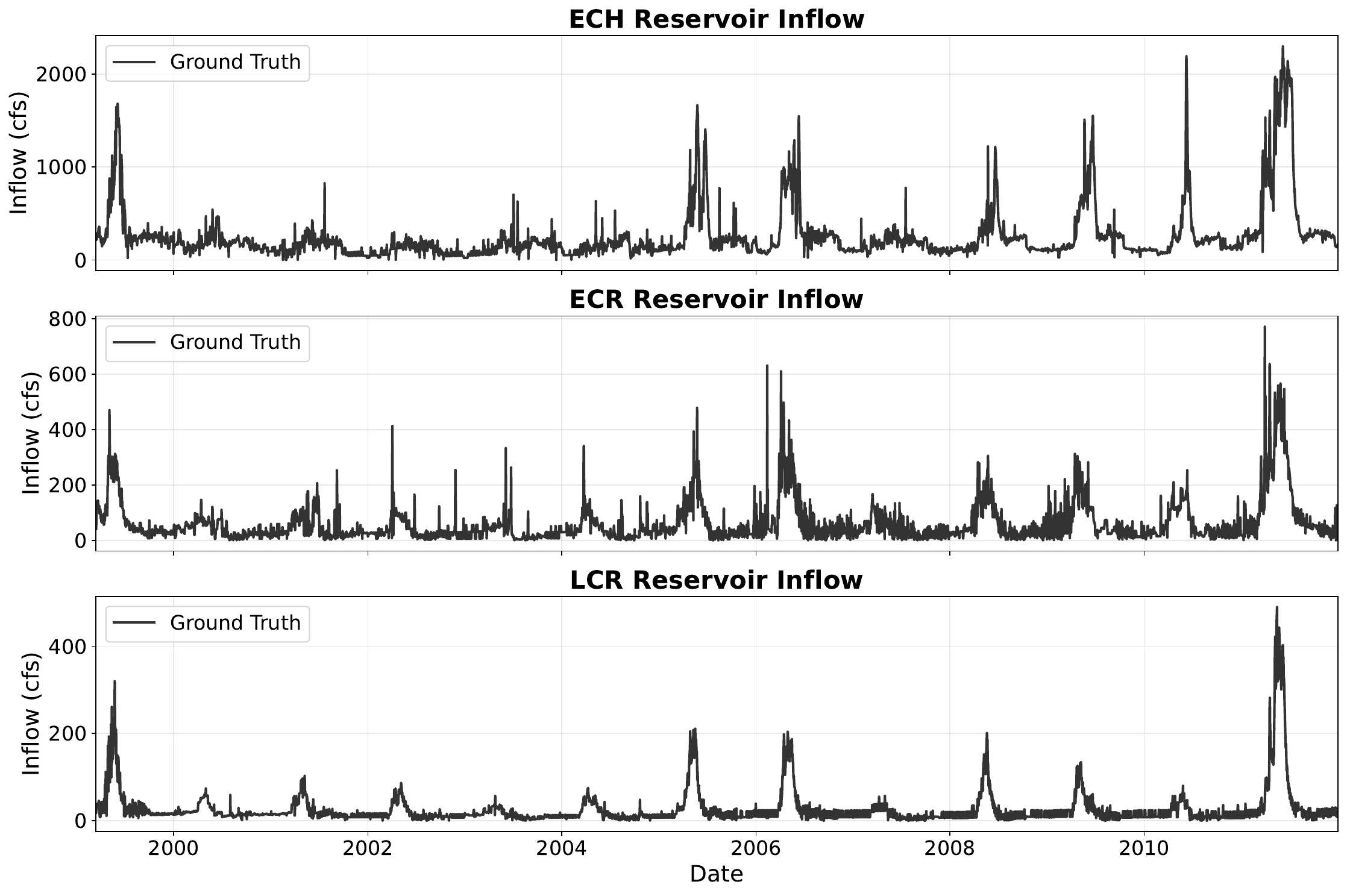}
    \caption{Reservoir inflow observations in reservoirs ECH, ECR, and LCR.}
    \label{fig:inflows_demo}
\end{figure}

\input{tabs/pretrain}
\begin{figure}[t]
    \centering
    \includegraphics[width=0.45\textwidth]{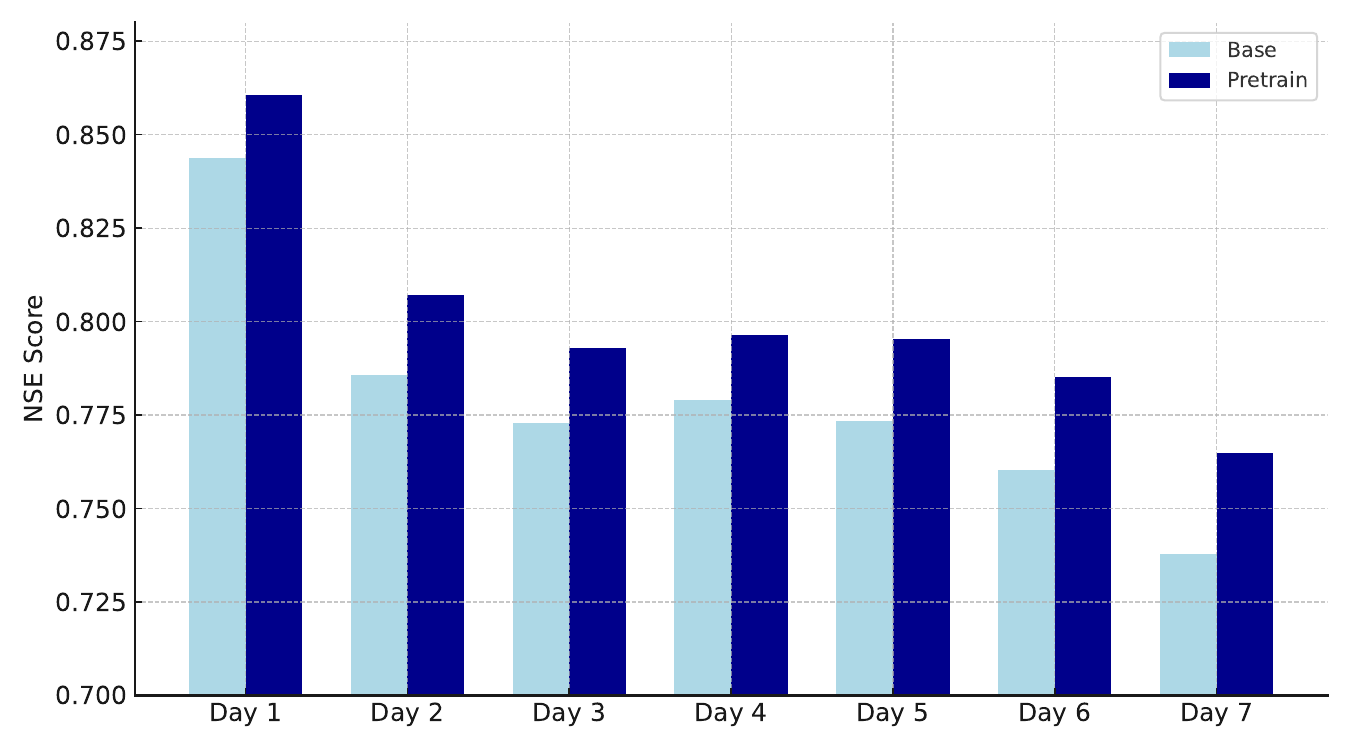}
    \caption{NSE Scores of inflow forecast in the reservoir HYR. ``Base'' means \model{} without pre-training, and ``Pretrain'' means the full \model{}.}
    \label{fig:pretrain_JOR}
\end{figure}

\subsection{Ablation on Pre-training}\label{ssec:ablation_pretrain}

Figure~\ref{fig:pretrain_JOR} shows the inflow forecasting of \model{} with and without pre-training on reservoir HYR, which has limited historical data (13 years) and is located at the lowest elevation, making it more susceptible to upstream influence. 
From the results, we observe consistent improvements in NSE scores across all forecast days when using pre-training. 
The performance gain is especially notable in longer horizons, especially on Days 6 and Day 7. 
For instance, pre-training improves the NSE score from $73.78\%$ to $76.48\%$ on Day 7.
This result highlights the effectiveness of pre-training in enhancing prediction accuracy, especially for data-scarce and topologically sensitive reservoirs.

\section{Conclusion}
We propose \model{} that adaptively learns spatial-temporal dependencies for multi-reservoir inflow forecasting. 
\model{} addresses two key challenges in GNN-based prediction: sensitivity to graph topology and the dynamic nature of inter-reservoir interactions. 
Through experiments on 30 reservoirs in the Upper Colorado River Basin, \model{} consistently outperforms baselines in both overall and reservoir-level evaluations. 
Further case studies reveal that \model{} effectively refines graph structures to reflect hydrological relevance, and that pre-training enhances performance in reservoirs with limited historical data. 
These findings suggest that adaptive graph learning offers a robust and generalizable approach for regional multi-reservoir inflow forecasting.

\section*{Acknowledgment}
This research was supported by the Seedling Project, funded by the U.S. Department of Energy (DOE) Water Power Technologies Office, and by Dan Lu’s Early Career Project, funded by the DOE Biological and Environmental Research Office. Additional support was provided through collaboration between the U.S. Air Force Life Cycle Management Center (LCMC) and Oak Ridge National Laboratory (ORNL). High-performance computing resources were provided by the Frontier system at the Oak Ridge Leadership Computing Facility, a DOE Office of Science User Facility.
This work was supported in part by the U.S. National Science Foundation under grants 2047843 and 2437621.

\bibliographystyle{IEEEtran}
\bibliography{main}

\end{document}

%% file: sections/intro.tex
Reservoir inflow prediction is crucial for effective water resource management and hydrological planning.
Accurate forecasting enables reservoir operators to optimize operations, ensuring adequate water and hydropower supply for domestic, agricultural, and industrial demands while maintaining ecological flows for environmental sustainability~\cite{latif2023review}. 
It is also essential for flood control strategies, allowing for proactive reservoir releases that mitigate downstream flooding risks~\cite{allawi2018review}. 
Consequently, researchers have developed various forecasting approaches, ranging from process-based hydrological models to data-driven quantitative methods.

Early inflow prediction methods mostly rely on processed-based hydrological models~\cite{fan2023explainable}.
Machine learning (ML) approaches have subsequently been adopted to enhance predictive accuracy, with ensemble methods like Random Forest showing improved performance in capturing associations between reservoir inflow and other environmental features~\cite{bernardes2022hydropower}. 
Deep learning (DL) techniques have also demonstrated superiority compared to conventional ML approaches, with convolution neural networks (CNNs) and recurrent neural networks (RNNs) achieving higher predictive accuracy.
Among these architectures, Encoder–Decoder Long Short-Term Memory (ED-LSTM) is commonly used for multi-step hydrologic prediction due to its strong temporal prediction capability~\cite{latif2023review, fan2025enhancing}.

However, existing ML models are mostly trained and evaluated on single-reservoir data, exploiting only temporal dynamics and disregarding spatial dependencies across reservoirs. 
These methods overlook the fact that reservoirs within the same watershed or river basin often exhibit correlated inflow patterns due to shared meteorological conditions and hydrological connectivity, which may cause performance degradation in large-scale management. 
To enable multi-reservoir learning, graph neural networks (GNNs) are adopted to make forecasting by data from both the target reservoir and neighboring reservoirs~\cite{deng2020cola, sun2021explore}.
Although delivering information along hydrological connections encoded in a river network can enhance predictions, there are two challenging problems when constructing the graph prior to training:

\begin{enumerate}
    \item \textbf{Topology Sensitivity.} The precision of most GNN-based approaches depends heavily on the graph quality. An overly dense graph may cause over-smoothing, whereas an overly sparse graph may hinder message passing. Consequently, the topology often requires manual tuning across different tasks or training data.
    \item \textbf{Dynamic Discrepancy.} Static graphs with graph convolution network (GCNs) are mostly adopted for prediction, but hydrological connectivity can change daily in real-world conditions. It might misrepresent the true interactions at many time steps, whenever the physical network differs from the predefined topology.
\end{enumerate}


To overcome these limitations, we propose \model{}, \uline{an end-to-end framework that learns adaptive, time-varying graphs reflecting evolving hydrological connectivity for multi-reservoir inflow forecasting}. Our approach constructs dynamic graphs where each reservoir is represented as a node with directed edges based on known hydrological connections.
A shared feature encoder and graph attention network are used to learn reservoir-specific representation based on neighboring reservoirs. 
Note that, we keep dynamic graphs across time steps by monitoring attention weights during the training phase, which allows the model to identify crucial edges that vary in importance depending on temporal and spatial conditions.
A transformer encoder-decoder then processes these dynamically adapted graphs to forecast future inflow. 
To fully utilize the whole reservoir data, we also design a semi-supervised paradigm for the pre-training phase to provide a robust initialization of \model{}.
Real-world reservoir data from the Upper Colorado River Basin are used for evaluation, which demonstrates \model{} improves the Nash–Sutcliffe Efficiency (NSE) metric~\cite{mccuen2006evaluation} by $2.0\%$ and $3.2\%$ over ED-LSTM and GCN+LSTM, which are the standard methods for spatial and temporal prediction. 
It also provides interpretable edge-level and time-step attention maps that align with known hydrological processes.
Our main contributions are:
\begin{enumerate}
    \item We firstly introduce adaptive graphs on transformer structure for multi-reservoir inflow forecasting, unifying spatial and temporal learning in an end-to-end network.
    \item The customized semi-supervised pretraining paradigm help graph-based approaches to fully utilize the whole reservoir records for the robust initialization.
    \item We provide built-in interpretability through edge-level and time-step attention scores, which highlights meaningful days and reservoirs for prediction.
\end{enumerate}

%% file: sections/related_work.tex
\subsection{Reservoir Inflow Forecasting in Hydrology}

There is a progression from ML to DL architectures for precise inflow predictions~\cite{azad2025developments}. 
Early ML approaches applied methods like support vector machines, decision-tree based models, and non-linear regression, often combined with wavelet analysis or bootstrap aggregation, to model reservoir inflows~\cite{allawi2018review, nourani2021prediction, latif2023review}. 
These models showed promise for short lead times, but struggled to capture the nonlinear dynamics of hydrological systems over longer periods~\cite{bernardes2022hydropower}. 
DL techniques like long short-term memory (LSTM) demonstrates superiority over hydrological models for short-term predictions~\cite{zhao2024comprehensive}.
However, a single LSTM might struggle with multi-step forecasts. 
Hence, ED-LSTM is then applied for sequence-to-sequence prediction of reservoir inflow~\cite{teegavarapu2025use, fan2023advancing, fan2023explainable}. 
Other than LSTMs, attention-based architectures and Transformers have emerged as promising approaches in the reservoir system, such as Dynamic Transformers\cite{xu2023dynamic} and temporal fusion transformers~\cite{muniz2025time}. 
These studies highlight encoder-decoder structures as powerful choices for multi-step hydrological forecasting.

Beyond pursuing high accuracy based on temporal changes on single reservoir, it is also important to get insights about hydrological interactions with other reservoirs in the shared river network. 
Therefore, developing neural networks leveraging inter-reservoir connections represents a promising direction for training regional rather than site-specific predictive models that utilize spatial knowledge beyond local information.

\subsection{Temporal Prediction of Graph Neural Network}

Graphs, consisting of sets of nodes and edges and their associated attributes, are fundamentally designed to represent unstructured data. 
Temporal GNNs have emerged as powerful tools for forecasting in applications such as transportation, navigation, and weather forecasting~\cite{yu2017spatio, zhou2024navigating, lam2023learning}. 
Similarly, recent work also leverages graphs to model watershed systems in the hydrology domain, where nodes correspond to basins and edges denote channels connecting them, enabling models to learn spatiotemporal dependencies based on hydrological variables~\cite{sun2021explore, taccari2024spatial, pan2025combining}.
Despite the growing adoption of GNNs across various domains, the predictive performance of these models remains highly sensitive to graph structure, leading researchers to focus on two critical aspects: 
(1) For graph construction, researchers invest considerable effort in ensuring graph accuracy, as incorrect edge connections can mislead model learning and introduce predictive errors~\cite{wu2020comprehensive, zhang2021graph, li2025implicit};
(2) For temporal dependencies, recent advances have moved beyond static graph representations toward dynamic frameworks capable of capturing time-varying node features and connectivity patterns~\cite{bui2022spatial, longa2023graph, chang2024dihan}.

Still, existing hydrological applications of GNNs have received limited attention on these fronts, with most studies adopting simplified graph structures without systematic consideration of graph quality or temporal dynamics.



%% file: tabs/notation.tex
\begin{table}[t!]
\centering
\caption{Summary of main symbols used in the model.}
\label{tab:notation}
\begin{tabular}{ll}
\toprule
Symbol & Meaning \\ \midrule
$N, |\mathcal{V}_t|$ & Number of reservoirs (nodes), here $N=30$ \\[2pt]

$F$ & Number of input features \\[2pt]

$T$ & Length of input records, here $T=30$ days \\[2pt]

$H$ & Forecast horizon, here $H=7$ days \\[2pt]

$X_i= \{\mathbf{x}_{i,t}\}^T_t$ & Input data of the $i$-th reservoir (on day $t$) \\[2pt]

$G_t, \mathcal{E}_t, \mathbf{A_t}$ & graph, edges, adjacency matrix on day $t$ \\[2pt]

$\mathcal{F}_\theta(\cdot), \mathcal{G}_\lambda(\cdot), \mathcal{T}_\omega(\cdot)$ & Different modules parameterized by $\theta, \lambda, \omega$ \\[2pt]

$d,m,r$ & Hidden dimensions in $\mathcal{F}_\theta, \mathcal{G}_\lambda,\mathcal{T}_\omega(\cdot)$ \\[2pt]

$p_i$ & The geographic position of reservoir $i$ \\[2pt]

$\mathcal{D}(\cdot)$ & The spatial distance between reservoir $i$ and $j$ \\[2pt]

$\mathbf{h}_{i,t}, \tilde{\mathbf{h}}_{i,t}\!\in\!\mathbb{R}^{d}$ & Embedding of reservoir $i$ on day $t$ in $\mathcal{G}_\lambda(\cdot)$ \\[2pt]

$\alpha_{ij,t}$ & Graph attention from node $j$ to $i$ on day $t$ \\[2pt]

$\beta_{k,\tau}$ & Temporal attention from step $\tau$ to step $k$ \\[2pt]

$\mathbf{z}_i\!\in\!\mathbb{R}^{d}$ & Reservoir-specific embedding after $\mathcal{T}_\omega(\cdot)$ \\[2pt]

$\hat{y}_{i,t+k}$ & Predicted inflow for reservoir $i$ at day $t+k$ \\[2pt]

$y_{i,t+k}$ & Observed inflow (ground truth) \\[2pt]
\bottomrule
\end{tabular}
\end{table}

%% file: sections/preliminary.tex
Following the setting in earlier studies~\cite{fan2023advancing, fan2023explainable}, the dataset $\mathcal{D} = \{X_1,X_2,\dots,X_N\}$ involves $N=30$ reservoirs in total. 
For the $i$-th reservoir, historical records $X_i=\bigl\{\mathbf{x}_{i,t-T+1},\dots,\mathbf{x}_{i,t}\bigr\} \in \mathbb{R}^{T \times F}$ include $F$ features (e.g, temperature, precipitation, and inflow) over the previous $T=30$ days, which serve as model input for temporal forecasting. 
Moreover, the targets are daily inflow forecasts $\mathbf{y}_{i,t+k}$ for the next $H=7$ days, where $k$ denotes the number of days ahead. 

To enable compound (i.e. multi-reservoir) forecasting, data on day $t$ can be also represented as a directed graph $G_t=(\mathcal{V}_t,\mathcal{E}_t)$ where $|\mathcal{V}_t|=N$ nodes correspond to connected reservoirs and edges are encoded by an adjacency matrix $\mathcal{E}_t:\mathbf{A_t}\in\{0,1\}^{N\times N}$ with binary values indicating the existence of hydraulic flow relationships. 
Note that, each observation record $\mathbf{x}_{i,t}$ is first scaled and then projected into a hidden representation  $\mathbf{h}_{i,t}$ in a $d$-dimensional embedding space through a feature extractor $\mathcal{F}_\theta(\cdot)$ with parameters $\theta$,  following the same paradigm as single reservoir training.  
The key difference is that the observation records $\mathbf{h}_{1,t},\mathbf{h}_{2,t},\dots,\mathbf{h}_{N,t}$ are then aggregated as node features $\mathcal{V}_t$ in $G_t$ and processed through the graph learning module $\mathcal{G}_\lambda(\cdot)$, where hydrological connectivity (e.g. spatial distance) can be learned from $\mathbf{A}_t$. 
Furthermore, the dimensions of hidden states in modules $\mathcal{F}_\theta(\cdot), \mathcal{G}_\lambda(\cdot), \mathcal{T}_\omega(\cdot)$ are denoted as $d, m, r$, respectively.
All notations are shown in Table \ref{tab:notation}.


%% file: sections/method.tex
We present an end-to-end model that turns historical 30 days of hydro-meteorological records from 30 reservoirs into seven-day inflow forecasts through four consecutive stages: 

\begin{enumerate}
    \item Historical observations of all reservoirs are projected by the feature extractor, where each reservoir has daily embeddings as node features across temporal graphs. 
    \item The graph learning module then removes unused edges in daily graphs initialized by spatial position to adjust node features by shared information. 
    \item Such enhanced representations are fed into an attention-based module, where we can capture temporal dependencies via the encoder-decoder structure to get reservoir-specific embeddings. 
    \item Given the embedding of each reservoir, we can then decode them and forecast future reservoir inflows.
\end{enumerate}

\subsection{Feature Extractor $\mathcal{F}_\theta(\cdot)$}

For each day $t$, the daily input data $\mathbf{x}_{t}\in\mathbb{R}^{N\times F}$ firstly goes through a feature extractor $\mathcal{F}_\theta(\cdot)$. 
In this paper, a multi-layer perceptron (MLP) are used to project raw features of $N$ reservoirs onto a $d$-dimensional vector space by
\begin{align}
    \mathbf{h}_{t}:=\operatorname{MLP}(\mathbf{x}_{t})\in\mathbb{R}^{N\times d}.
\end{align}
Here each reservoir $i$ get daily embedding $\mathbf{h}_i = \{\mathbf{h}_{i,t}\}_{t=1}^{N}$ for every time step (i.e. day) $t$. 
Note that, daily embeddings across $N$ reservoirs on the dataset should be fed together as a sequence $\{\mathbf{h}_{1,t}, \mathbf{h}_{2,t}, \dots, \mathbf{h}_{N,t}\}$ for graph learning.

\vspace{0.05in}
\noindent
\textbf{Initialize $\theta$ by Pretraining.} 
For multi-reservoir prediction, a critical challenge is how to effectively utilize historical data across all reservoirs.  
Temporal alignment represents a prerequisite step, as real-world datasets inevitably contain daily records with missing values and mismatched dates across different reservoirs. 
Previous approaches either employ mask tokens for missing values or directly discard mismatched dates, both of which fail to fully exploit available data. 
To address this limitation, we design a pretraining paradigm that specifically leverages these mismatched and previously unused records to provide better initialization for $\theta$ in $\mathcal{F}_\theta(\cdot)$, which enables $\mathcal{G}_\lambda(\cdot)$ to extract more precise node features.
To learn robust reservoir representations while maintaining prediction accuracy, a semi-supervised learning that combines both contrastive and supervised losses is employed. 

For the contrastive term, We draw $b$ samples from each reservoir, stack them into $\{\mathbf{x}_i\}^{B}$ with $B=bN$ in every batch. 
We then treat two embeddings originating from that reservoir as a positive pair and all other embeddings as negatives. 
Each reservoir in negative set maintains a momentum-updated prototype $\mathbf{c}_{i}\in\mathbb{R}^{d}$, and positive ones receive mild augmentations (i.e. Gaussian noise) on both temperature and precipitation to prevent trivial matching. 
Here, we adopt InfoNCE~\cite{oord2018representation} as the contrastive objective to maximize the similarity between positive pairs while minimizing similarity to negative pairs in the batch, as in Eq. (\ref{eq:contrastive_loss}),
\begin{equation}
\mathcal{L}_{\text{c}}=
-\log
\frac{\exp\bigl(\langle\mathbf{h}_i,\mathbf{h}_{i}^{+}\rangle/\tau\bigr)}
{\sum_{s\in\mathcal{S}}\exp\bigl(\langle\mathbf{h}_{i},
\mathbf{h}_{s}^{*}\rangle/\tau\bigr)}, \quad *\in \{-,+\}
\label{eq:contrastive_loss}
\end{equation}
where $\langle\cdot, \cdot\rangle$ denotes the cosine similarity, $\tau$ is the temperature, and $\mathcal{S}$ is the entire set of contrastive samples. 
Note that $\mathcal{L}_c$ should be summed over all reservoirs.

Meanwhile, the daily embedding of reservoir $i$ feed an extra classifier with weights $\mathbf{w}_\psi$ supervised by the ground truth inflow $\mathbf{y}_i$ via mean-squared error (MSE), as in Eq. (\ref{eq:mse_loss}), 
\begin{align}
    \mathcal{L}_{s}=
    \text{MSE}(\hat{\mathbf{y}}_{i,t+k}, \mathbf{y}_{i,t+k})=
    \sum_{k\leq H}\|\mathbf{w}_\psi \bar{\mathbf{h}}_{i}-\mathbf{y}_i\|^{2}.
    \label{eq:mse_loss}
\end{align}
where $\bar{\mathbf{h}}_{i}=\frac{1}{T}\sum_T\mathbf{h}_{i,t}$ over daily records on all $T=30$ days.
The total objective is a weighted sum $\mathcal{L}_p=(1-\alpha)\mathcal{L}_{c}+\mathcal{L}_{s}$, and the hyperparameter $\alpha$ can be manually tuned. 
After pretraining $\mathcal{F}_\theta(\cdot)$, the feature extractor can shape a reservoir-aware feature space, which also help \model{} accelerate convergence and yield robust predictions.

\subsection{Adaptive Graph Learning $\mathcal{G}_\lambda(\cdot)$}

Here we aim to augment learned representations by external spatial information beyond the training dataset. 
By considering the limitations of both graph accuracy and temporal dependencies, we construct $\mathcal{G}_\lambda(\cdot)$ in three steps.

\subsubsection{Graph Construction}
The spatial position of each reservoir $i$ is identified by its geographic coordinate $\mathbf{p}_{i}=(\varphi_{i},\lambda_{i})$ and surface elevation $h_{i}$, so that the distance between reservoir $i$ and $j$ can be measured by 
  
\begin{align}
d_{ij}=\mathcal{H}\bigl(\mathbf{p}_{i},\mathbf{p}_{j}\bigr),\qquad    
\delta_{ij}=\begin{cases}
1,&h_{j}<h_{i}\\[2pt]
0,&\text{otherwise}.
\end{cases}
\end{align}
Here $\mathcal{H}(\cdot)$ is calculated by the Haversine distance~\cite{qi2022regionalization} commonly used in the water-resource management, and the directed edge $\delta_{ij}$ from reservoir $i$ to $j$ is allowed only from an upstream reservoir to a downstream one when $h_{j}<h_{i}$. 
Specifically, every reservoir $i$ firstly select top $k$-th nearest reservoirs into the candidate set $\mathcal{C}_{i}$ with $k=|\mathcal{C}_{i}|$, and then filters out those that are not in the downstream of reservoir $i$:
\begin{align}
\mathcal{N}_{i} =\{j\in\mathcal{C}_{i}^{\text{dist}}:\delta_{ij}=1\}.
\end{align}
$\mathcal{E}=\{(i,j):j\in\mathcal{N}_{i}\}$ is the set of edge indices where $\mathcal{N}_{i}$ is chosen from $\mathcal{C}_{i}$. 
The adjacency matrix $\mathbf{A}\in\{0,1\}^{N\times N}$ can be then defined by $a_{ij}=1$ when $\delta_{i,j}$ exists and 0 otherwise. 
Note that $\mathbf{A}$ is also the edge initialization across all temporal graphs $G_1,G_2,\dots,G_T$.

\subsubsection{Graph Attention Network}
After initializing daily graph $G_t$ by the static spatial connectivity $\mathbf{A}$ with $t=1,2,\dots,T$, a GNN module will be used to further adjust daily embedding $\mathbf{h}_{i,t}$ for each reservoir by considering the nearby-reservoir information. 
Here, spatial information is propagated through two-layer Graph Attention Network (GAT), which means the adjusted embedding $\tilde{\mathbf{h}}_{i,t}$ will consider its 2-hop neighboring reservoirs.
For instance, we can get $\tilde{\mathbf{h}}_{i,t}$ through the first layer, as in Eq. (\ref{eq:gat}),
\begin{align}
e_{ij,t}&= \text{LeakyReLU}\bigl(\mathbf{a}^{\top}[\mathbf{w}\mathbf{h}_{i,t}\Vert\mathbf{w}\mathbf{h}_{j,t}]\bigr), \notag \\ 
\alpha_{ij,t}&= \frac{\exp(e_{ij,t})}{\sum_{j'\in\mathcal{N}_{i}} \exp(e_{ij',t})} \in \mathbb{R}, \notag \\
\tilde{\mathbf{h}}_{i,t}&= \sigma\!\bigl(\frac{1}{K}\sum_{k=1}^{K}\sum_{j\in\mathcal{N}_{i}}\alpha^{(k)}_{ij,t}\cdot\mathbf{w}^{(k)}\mathbf{h}_{j,t}\bigr),
\label{eq:gat}
\end{align}
where $K$ is the number of heads, $\mathbf{w}^{(k)}\in\mathbb{R}^{d\times d}$ and $\mathbf{a}\in\mathbb{R}^{2d}$ are learnable, $[\cdot\Vert\cdot]$ denotes the simple concatenation over two vectors, $\mathcal{N}_{i}$ downstream neighbors of $i$, and $\sigma$ is the ReLU activation function. 
Similarly, the second GAT layer further updates $\tilde{\mathbf{h}}_{i,t}$ by $\text{GATConv}(\tilde{\mathbf{h}}_{i,t})$, and the daily node feature matrix $\mathbf{H}_{t}\in\mathbb{R}^{N\times d}$ is then forwarded to $\mathcal{T}_\omega(\cdot)$.

\subsubsection{Dynamic Adaptation on Temporal Graphs} 
Although temporal change among daily node features can be learned by following the architecture of $\text{GATConv}(\tilde{\mathbf{h}}_{i,t})$, temporal graphs $G_t$ still remain unchanged with the same $\mathbf{A}$. 
We cannot simplify the potential connectivity by static edges, which ignores the evolution fact that hydraulic edges among reservoir might change even on every single day. 
Therefore, we focus on $\alpha_{ij,t}$, the important parameters in the GAT module.

Specifically, these attention weights serve as internal measures of hydraulic influence.  
The coefficient produced by certain head $h$ in layer $\ell\in\{1,2\}$ for edge $(i,j)$ on day $t$ is denoted as $\alpha_{ij,t}^{(\ell,h)}$. 
We condense these scores into 
\begin{equation}
\bar{\alpha}_{ij,t}=\frac{1}{2K}\sum_{\ell=1}^{2}\sum_{k=1}^{K}\alpha_{ij,t}^{(\ell,k)}.
\end{equation}
During each training epoch we accumulate the expectation over the $T$ input snapshots $\tilde{\alpha}_{ij}=\mathbb{E}[\bar{\alpha}_{ij,t}]^{T}_{t=1}$, which represents the average usefulness of edge $(i,j)$ for the current mini-batch. 
After every $m$ epochs the model pauses, and edges whose running mean falls below a fixed threshold $\tau$ are masked out.
For instance, we set $\delta_{i,j}=0$ in $G_t$ when $\tilde{\alpha}_{ij}<\tau$ for day $t$, and vice versa.
Such pruning approach on graphs not only help \model{} adjust daily graphs automatically, but also enhance robustness of prediction adaptively.
Moreover, it sharpens spatial focus, reduces over-smoothing, and supplies a self-selected river network for downstream analysis.

\subsection{Temporal Encoder–Decoder Learning $\mathcal{T}_\omega(\cdot)$}

Following the setting of other sequence-to-sequence-based models, we aim to let $\mathcal{T_\omega(\cdot)}$ learn temporal patterns based on the sequence of adjusted embeddings $\tilde{\mathbf{H}_i} = [\tilde{\mathbf{h}}_{1},\tilde{\mathbf{h}}_{2},\dots,\tilde{\mathbf{h}}_{T}]$, which can be stacked into $\mathbf{z}_i\in\mathbb{R}^{d}$ across $T$ days. 
We adopt Transformer rather than ED-LSTM as $\mathcal{T_\omega(\cdot)}$ for temporal module.
There are two reason: 
(1) A powerful tool is required for more complicated patterns after dynamic graphs;
(2) Attention scores can help us focus more on specific dates with interpretability.
Here we explain the procedure by the encoder $\mathcal{T}_e$ and decoder $\mathcal{T}_d$ individually. 

For $\mathcal{T}_e$, the position vector $\mathbf{P}\in\mathbb{R}^{T\times d}$ across time steps is added to $\tilde{\mathbf{H}_i}$ of every reservoirs $\mathbf{E}_{i}=\tilde{\mathbf{H}_i}+\mathbf{P}$.
The input then passes through the encoder, as in Eq. (\ref{eq:seq_encoder}),
\begin{equation}
    \mathbf{M_i}=\mathcal{T}_{e}(\mathbf{E}_{i})=
    \mathcal{T}_{e}(\tilde{\mathbf{H}_i}+\mathbf{P})
    \in\mathbb{R}^{T\times d},
    \label{eq:seq_encoder}
\end{equation}
where the contextualized output $\mathbf{M_i}$ then serves as the memory for the decoder via cross-attention parameters.

Given the encoder memory $\mathbf{M}_{i}$, the
decoder can provide the final state before prediction, as in Eq. (\ref{eq:seq_decoder}),
\begin{equation}
    \mathbf{z}_{i} =
    \mathcal{T}_{d}(
        \mathbf{M}_{i})\in\mathbb{R}^{d},
    \label{eq:seq_decoder}
\end{equation}
where $\mathbf{z}_{i}\in\mathbb{R}^{d}$ is the hidden state vector, and $\mathcal{T}_{d}(\cdot,\cdot)$ involves a masked self‑attention layer, cross‑attention with $\mathbf{M}_{i}$, and a linear
projection. 
Future predictions can be obtained by recursively feeding previous states $\mathbf{z}_{i}$ back.

\subsection{Training \& Inference}

A single classifier with learnable weights $\mathbf{w}_d$ is then used as the predictor.
The future inflow $k$-day ahead can be calculated by $\hat{y}_{i,t+k} = \mathbf{w}_d \cdot \mathbf{z}_i$. 
Similar to the pretraining phase, MSE loss is used for the training stage, as in Eq. (\ref{eq:mse_full}),
\begin{equation}
    \mathcal{L}
      =\frac{1}{NH}
        \sum_{i=1}^{N}\sum_{k=1}^{H}
        \bigl(\hat{y}_{i,t+k}-y_{i,t+k}\bigr)^{2},
    \label{eq:mse_full}
\end{equation}
aggregated over all $N$ reservoirs and $H$ leading times. 
Dynamic graph tuning is interleaved with learning: running edge-attention means are evaluated and edges with mean weight below the threshold \(\tau\) are masked every $m$ epochs. 
During inference the final pruned topology is fixed, the decoder rolls out forecasts, and the model outputs the seven-day inflow vector for all thirty reservoirs in a single forward pass.

%% file: tabs/data_stats.tex
\begin{table}[t]
    \centering
    \caption{Information of the 30 reservoirs analyzed in this study.}
    \label{tab:stats}
    \resizebox{0.47\textwidth}{!}{%
    \begin{tabular}{llcc}
    \hline
    \textbf{Initials} & \textbf{Names} & \textbf{Data start year} & \textbf{Data length (years)} \\
    \hline
    BSR & Big Sandy Reservoir & 1990 & 22 \\
    CAU & Causey Reservoir & 1999 & 13 \\
    CRY & Crystal Reservoir & 1982 & 30 \\
    DCR & Deer Creek Reservoir & 1987 & 25 \\
    DIL & Dillon Reservoir & 1985 & 27 \\
    ECH & Echo Reservoir & 1982 & 30 \\
    ECR & East Canyon Reservoir & 1992 & 20 \\
    FGR & Flaming Gorge Reservoir & 1982 & 30 \\
    FON & Fontenelle Reservoir & 1990 & 22 \\
    GMR & Green Mountain Reservoir & 1982 & 30 \\
    HYR & Hyrum Reservoir & 1999 & 13 \\
    JOR & Jordanelle Reservoir & 1997 & 15 \\
    JVR & Joes Valley Reservoir & 1996 & 16 \\
    LCR & Lost Creek Reservoir & 1998 & 14 \\
    LEM & Lemon Reservoir & 1982 & 30 \\
    MCP & Mcphee Reservoir & 1991 & 21 \\
    MCR & Meeks Cabin Reservoir & 1998 & 14 \\
    NAV & Navajo Reservoir & 1986 & 26 \\
    PIN & Pineview Reservoir & 1990 & 22 \\
    RFR & Red Fleet Reservoir & 1989 & 23 \\
    RID & Ridgway Reservoir & 1990 & 22 \\
    ROC & Rockport Reservoir & 1982 & 30 \\
    RUE & Ruedi Reservoir & 1982 & 30 \\
    SCO & Scofield Reservoir & 1996 & 16 \\
    SJR & Silver Jack Reservoir & 1992 & 20 \\
    STA & Starvation Reservoir & 1982 & 30 \\
    STE & Steinaker Reservoir & 1982 & 30 \\
    TPR & Taylor Park Reservoir & 1982 & 30 \\
    USR & Upper Stillwater Reservoir & 1991 & 21 \\
    VAL & Vallecito Reservoir & 1986 & 26 \\
    \hline
    \end{tabular}%
    }
\end{table}

%% file: tabs/main_results.tex
\begin{table*}[t]
\centering
\caption{Overall and per–day NSE (\%) for all baselines. Numbers in parentheses show the standard deviation over five independent runs.}
\label{tab:overall_results}
\begin{tabular}{lcccccccc}
\toprule
Model & Overall & Day\,1 & Day\,2 & Day\,3 & Day\,4 & Day\,5 & Day\,6 & Day\,7\\
\midrule
ED--LSTM      & 89.47\,(1.65) & 95.68\,(0.94) & 93.87\,(2.96) & 91.49\,(1.12) & 89.27\,(0.87) & 87.10\,(1.01) & 85.27\,(0.60) & 83.63\,(0.67)\\
Transformer   & 87.90\,(0.43) & 91.70\,(2.16) & 91.61\,(1.91) & 89.51\,(2.44) & 88.12\,(1.21) & 86.18\,(2.71) & 85.02\,(1.71) & 83.20\,(2.28)\\
GCN+LSTM      & 88.29\,(0.46) & 94.09\,(2.82) & 92.28\,(2.75) & 90.18\,(1.65) & 88.20\,(1.17) & 86.20\,(0.35) & 84.40\,(2.88) & 82.66\,(1.99)\\
\model{}       & \textbf{91.45}\,(2.60) & \textbf{96.20}\,(1.12) & \textbf{95.05}\,(1.88) & \textbf{93.31}\,(2.56) & \textbf{91.76}\,(1.95) & \textbf{89.78}\,(0.95) & \textbf{88.03}\,(0.38) & \textbf{86.00}\,(2.77)\\
\bottomrule
\end{tabular}
\end{table*}